\documentclass[conference]{IEEEtran}

\usepackage{algpseudocode}    
\usepackage{algorithm}
\usepackage{threeparttable}
\usepackage{amsmath}
\usepackage{amssymb}
\usepackage{amsfonts}
\usepackage{booktabs}
\usepackage{multirow}
\usepackage{subfigure}
\usepackage{graphicx}         
\usepackage{color}
\usepackage{cite}             
\usepackage{comment}          
\usepackage{soul}             
\soulregister\cite7
\soulregister\ref7
\soulregister\pageref7
\usepackage{amsthm}
\usepackage{etoolbox}         
\usepackage{url}
\usepackage{nth}              
\usepackage{bm}               
\usepackage{cleveref}

\makeatletter
\let\OldStatex\Statex
\renewcommand{\Statex}[1][3]{%
  \setlength\@tempdima{\algorithmicindent}%
  \OldStatex\hskip\dimexpr#1\@tempdima\relax
}
\makeatother

\graphicspath{{figs/}}
\usepackage[auth-lg]{authblk}
\usepackage{amsmath}
\usepackage{multirow}
\usepackage{colortbl}
\usepackage{comment}
\usepackage{dblfloatfix}

\begin{document}

\title{
On Intrinsic Dataset Properties for Adversarial Machine Learning
}

\IEEEoverridecommandlockouts
\author[1]{Jeffrey Z. Pan}
\author[1]{Nicholas Zufelt}

\affil[1]{Phillips Academy, Andover, MA 01810}
\affil[ ]{\{jpan21, nzufelt\}@andover.edu}

\thispagestyle{empty} 
\maketitle

\begin{abstract}

Deep neural networks (DNNs) have played a key role in a wide range of machine learning applications. However, DNN classifiers are vulnerable to human-imperceptible adversarial perturbations, which can cause them to misclassify inputs with high confidence. Thus, creating robust DNNs which can defend against malicious examples is critical in applications where security plays a major role. In this paper, we study the effect of intrinsic dataset properties on the performance of adversarial attack and defense methods, testing on five popular image classification datasets — MNIST, Fashion-MNIST, CIFAR10/CIFAR100, and ImageNet. We find that input size and image contrast play key roles in attack and defense success. Our discoveries highlight that dataset design and data preprocessing steps are important to boost the adversarial robustness of DNNs. To our best knowledge, this is the first comprehensive work that studies the effect of intrinsic dataset properties on adversarial machine learning. 

\end{abstract}

\section{Introduction}
\label{sec:intro}


Deep learning has seen tremendous success recently in many applications including image and voice recognition, text generation and classification, and autonomous driving. However, researchers have shown that deep neural networks (DNNs) are vulnerable to adversarially perturbed inputs \cite{szegedy2013intriguing}. Thus, much work has been done developing strategies to create models robust to such adversarial inputs. 
Most current literature in the field of adversarial machine learning focuses primarily on image recognition tasks. There are also recent works on the applicability of adversarial attacks on other tasks such as speech \cite{carlinivoicecommands}\cite{Zhang_2017}, text \cite{jin2019bert}, and reinforcement learning \cite{s2017adversarial}. 

As machine learning and DNNs become more prevalent across a wide variety of application fields, including sensitive applications that require security and privacy, creating robust DNNs that perform accurately in spite of adversarially perturbed inputs becomes a critical task for deploying such networks.

Although much work has been previously done exploring new adversarial attacks for benchmarking adversarial robustness and new defense methods to secure networks \cite{yuan2019adversarial}\cite{ren2020adversarial}, no work to our best knowledge has explored the effect of dataset properties on adversarial attack and defense performance. Given that proper dataset design and preprocessing plays a crucial role in ensuring the accuracy of DNNs, we believe that such dataset properties may also affect the performance of adversarial attack and defense methods. We suggest intrinsic dataset properties, such as image size and image contrast, may impact adversarial robustness and experimentally demonstrate the effect of these properties against a variety of attack and defense methods on five popular image classification datasets.

The main contributions of this paper are  as follows.

\begin{itemize}
\item We first suggest and discuss two intrinsic dataset properties, image size and image contrast, which may affect adversarial robustness.

\item We explore the effects of these dataset properties on adversarial robustness, showing that they affect attack and defense performance across several popular datasets and under different attack and defense methods.

\item We give explanations for why these dataset properties may affect adversarial robustness.

\item We discuss possible applications of leveraging these properties for future robust dataset design, preprocessing steps, and adversarial machine learning.
\end{itemize}

The rest of the paper is organized as follows.
Section~\ref{sec:background} provides preliminaries of neural networks and adversarial machine learning, and then explains the adversarial attacks and defenses we will use in our study.
Section~\ref{sec:properties} explains the intrinsic dataset properties such as image size and image contrast that we believe will play important roles in adversarial machine learning. Section~\ref{sec:experiments} shows extensive experimental results on intrinsic dataset properties on adversarial robustness, using various attack and defense techniques on five popular datasets: MNIST, Fashion-MNIST, CIFAR10, CIFAR100, and ImageNet.
Section~\ref{sec:conclusion} concludes the paper with future research directions and potential applications.
\section{Preliminaries}
\label{sec:background}

In this section, we first provide some background information about deep 
learning and neural networks, particularly convolutional neural networks. Then, we will provide a brief summary of adversarial machine learning, and different attack and defense techniques.

\subsection{Deep Learning and Neural Networks}
\label{intro:ml}

The widespread use of deep learning techniques and neural networks began with AlexNet \cite{alexnet}, a convolution neural network that achieved state-of-the-art results on several image classification benchmarks in 2012. Neural networks allow machine learning models to learn features from raw input data. The intermediate layers, known as hidden layers, of a neural network help it extract underlying patterns within the raw, unstructured data, allowing for better model performance. Thus, DNN architectures have found incredible success in learning from complex, multidimensional inputs. 

Convolutional neural networks (CNNs) in particular have found widespread use in image recognition tasks. A CNN is mainly comprised of convolutional layers, which attempt to extract useful features from two-dimensional input data to create a feature map. Then, the model uses pooling and downsampling layers to reduce the dimensionality of intermediate feature maps, retaining only the most important features. Finally, a fully-connected layer predicts a final result from the previously-computed feature map. CNNs thus can take advantage of the intrinsic structure of two-dimensional data, leading to their excellent performance and popularity on image recognition tasks.

\subsection{Adversarial Machine Learning}
\label{intro:adversarial}

The goal of adversarial machine learning is to study methods of exploiting machine learning pipelines to identify weaknesses and build more robust machine learning models. Thus, it is imperative to first understand the different types of adversarial attacks.
Adversarial attacks against neural networks fall into several categories \cite{Biggio_2014} \cite{yuan2019adversarial}:

\textbf{Evasion.} Evasion attacks aim to evade a machine learning system by passing it carefully crafted inputs at inference time. If successful, an evasion attack would craft malicious inputs that cause the machine learning system to fail to produce correct outputs. Evasion attacks are the most common adversarial attack and thus the most-studied type of adversarial attack.

\textbf{Poisoning.} Poisoning attacks occur during the training phase of a machine learning system. In a poisoning attack, an attacker attempts to insert malicious inputs into the training data of a model with the goal of compromising the model's training.

\textbf{Extraction.} Extraction attacks attempt to extract the details of a model, its training procedure, and training data. With the application of machine learning in sensitive fields, extraction attacks could allow malicious agents to access proprietary models and sensitive data such as medical records.

In this paper, we primarily focus on evasion attacks as they comprise the majority of adversarial attacks. Evasion attacks fall into several categories based on the adversary's access to the targeted model and training data:

\textbf{White-Box Attack.} In a white-box attack, we assume that the adversary has full access to the model parameters, training algorithm, and training data distribution. We focus specifically on the white-box threat model in this paper.

\textbf{Gray-Box Attack.}
In a gray-box attack, we assume that the adversary knows about the target model and the defense, but not the parameters of the defense. The adversary only knows the underlying learning algorithm and model topology but has no access to the training data or the trained parameters.

\textbf{Black-Box Attack.} Under a black-box threat model, the adversary only has query access to the targeted model, meaning they can only access the prediction and confidence score for a given input.

As most popular attack methods in the literature are gradient-based and thus fall under the white-box threat model, we operate under the white-box threat model for all of our experiments, giving the attacker full access to the model and gradients. By using attack methods that work in the white-box threat model, we hope to make it easier to compare our results to the existing literature.

\subsection{Adversarial Attacks.}

We now provide a brief overview of each attack method used in our experiments.

\subsubsection{Fast Gradient Sign Method (FGSM)}
\label{evasion:fgsm}

Goodfellow et al. \cite{goodfellow2014explaining} introduced the Fast Gradient Sign Method (FGSM), which quickly takes the gradient of the cost function with respect to a given input. For a maximum allowable perturbation $\epsilon$, a cost function $J$, and a model with parameters $\theta$, input $x$, and label $y$, FGSM linearizes the cost function at the current $\theta$ and returns a perturbation $\rho$ as a step of size $\epsilon$ in the direction that maximizes the cost function:
$$\rho = \epsilon \cdot \text{sign}(\nabla_x J(\theta, x, y))$$

As a simple gradient calculation that can be computed quickly through backpropagation, FGSM provides a cost-effective method to adversarially perturb an input to a neural network.

\subsubsection{Basic Iterative Method (BIM)}

An extension of FGSM, the Basic Iterative Method (BIM) \cite{kurakin2016adversarial} applies FGSM multiple times with a small step size, clipping pixel values after each step to ensure that the perturbed image remains within the $\epsilon$-neighborhood of the original image. Formally, the perturbed image is defined as:
$$x_0^{adv} = x, x_{N+1}^{adv} = \text{Clip}_{x, \epsilon}\{x_N^{adv} + \alpha \cdot \text{sign}(\nabla_x J(\theta, x_N^{adv}, y)\}.$$

\noindent where $\alpha$ is the step size for each iteration of BIM.

\subsubsection{Projected Gradient Descent (PGD)}
\label{evasion:pgd}

Projected Gradient Descent (PGD) is a first-order attack method introduced by Madry et al. \cite{madry2017deep}. PGD is a multi-step, iterative gradient-based attack that attempts to maximize the cost function via the following equation:
$$x^{t+1} = \Pi_{x+S}(x^t + \epsilon \cdot \text{sign}(\nabla_x J(\theta, x, y))$$

\noindent where $\Pi$ is the projection onto the ball of the maximum allowable perturbation $x+S$. Essentially, one can view PGD as a multi-step variant of FGSM. The additional attack effectiveness from multiple iterations makes PGD a powerful first-order adversary.

\subsubsection{DeepFool}
\label{evasion:deepfool}

DeepFool \cite{moosavidezfooli2015deepfool} attempts to calculate adversarial perturbations by perturbing a given input $x_0$ along the orthogonal projection of $x_0$ to its separating decision boundary. At each iteration, DeepFool linearizes the model function $f$ at the current point $x_i$ and then computes the minimal perturbation $r_i$ as:
$$\underset{r_i}{\text{argmax}} {||r_i||}_2 \text{ subject to }  f(x_i) + \nabla f(x_i)^{T}r_i = 0.$$

\subsection{Adversarial Defense Strategies}

Current adversarial defense strategies fall into three main categories — compression-based defenses, adversarial retraining, and modeling-based defenses. Compression-based defenses aim to reduce the impact of adversarial perturbations by attempting to smooth such perturbations, thus reducing their effectiveness. Dziugaite et al. \cite{dziugaite2016study} explored the use of JPEG compression to smooth input images before model inference, discovering that JPEG compression is effective in some cases in reducing the drop in classification accuracy for small perturbation sizes. Additionally, Guo et al. \cite{guo2017countering} attempted using total variance minimization (TVM) \cite{rudintvm} and other image transformations to counter adversarial perturbations, discovering that total variance minimization can significantly improve the robustness of DNNs to adversarial attack.

Adversarial retraining attempts to retrain the original model on adversarially-generated inputs to improve its robustness. However, retraining often fails to generalize across different attack methods, or even the same attack method run with different parameters. Several works have proposed different training schemes aimed at improving the generalizability of adversarial retraining. Madry et al. \cite{madry2017deep} suggested retraining using adversarial examples generated using PGD, as retraining on more powerful adversarial examples often leads to model robustness against weaker examples. Tramèr et al. \cite{tramer2017ensemble} proposed ensemble adversarial training, which augments a model's training data with perturbations transferred from other models. Finally, modeling-based approaches try to use a separate network architecture to detect adversarial examples, either discarding them or attempting to remove adversarial perturbations. Song et al. \cite{song2017pixeldefend} proposed PixelDefend, a modeling-based defense that uses the log likelihoods of a PixelCNN \cite{oord2016pixel} model to detect adversarial examples, as such images lie outside of the probability distribution of the clean training data. They then purify adversarial examples using the PixelCNN, modifying the perturbed image so that it can be classified correctly. 

As the latter two categories of defense techniques are not model and attack agnostic, and our goal is to study the effect of intrinsic dataset properties across several attacks and datasets, we mainly focus on two compression defenses (TVM and JPEG) in this work. Being model and attack agnostic, such compression-based defenses provide a baseline to measure the relationship between intrinsic dataset properties and defense performances. 
\section{Intrinsic Dataset Properties}
\label{sec:properties}

In this section, we will discuss some intrinsic properties of datasets, formulate metrics to quantify them, and hypothesize how such properties might affect adversarial attack and defense performance.

\subsection{Image Size}

\begin{figure}[h]
\centering
\includegraphics[scale=0.5]{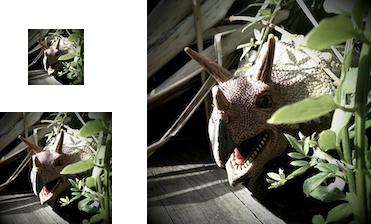} \\
\caption{
Three different resized versions ($56 \times 56$, $112 \times 112$, and $224 \times 224$) of an image taken from ImageNet.}
\label{fig:size_example} 
\end{figure}

We first consider the size of input images as an intrinsic dataset property. Fig.~\ref{fig:size_example} shows an image taken from the ImageNet dataset resized to three different crops. The top left image is cropped to $56 \times 56$, the bottom left image is cropped to $112 \times 112$, and the right image is cropped to $224 \times 224$ which is the standard crop used for many ImageNet classification networks. Current neural networks work best with fixed image sizes, as a trained network's weights will only perform optimally for features of a certain predetermined scale. Although to the best of our knowledge no comprehensive study has been done on the effect of input image size on neural network performance, intuitively a larger input size should result in higher classification accuracy, as a larger input image means more input features that a network could use for prediction. However, larger input images usually come at a higher computational cost since models are usually deeper to represent functions that fit this higher dimensional input. Additionally, simply storing and loading larger input images increases memory usage, thus making training with very high quality images infeasible for many applications, particularly for edge applications with limited resources. We discover, however, that larger input images may be easier to adversarially attack. Thus, such larger images may not be optimal for adversarial robustness, as we shall see in \ref{sec:experiments}.

\begin{figure*}[h!]
\centering
\includegraphics[scale=0.25]{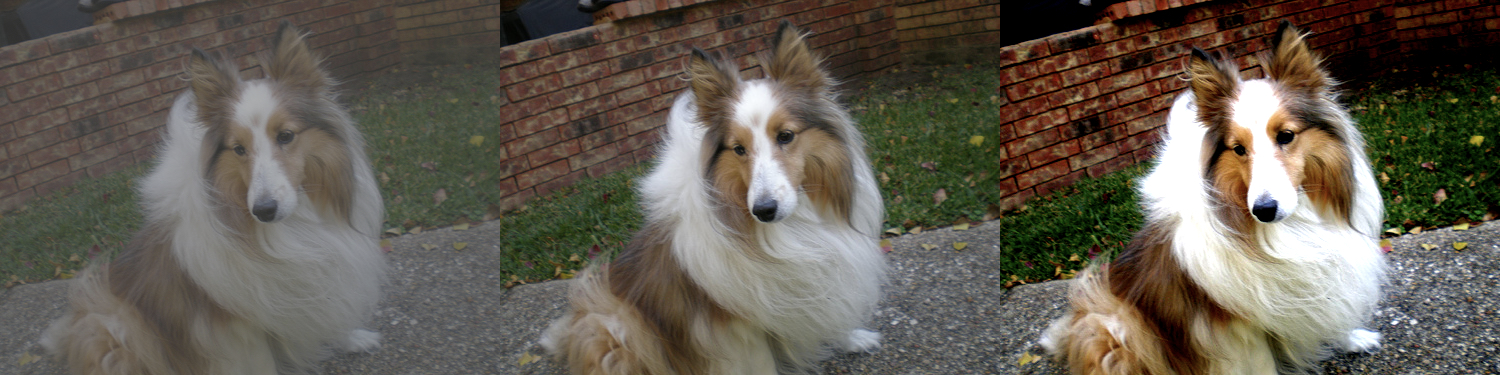} \\
\caption{
An ImageNet image at three different contrast settings.}
\label{fig:contrast_example} 
\end{figure*}

\subsection{Image Contrast}

Image contrast is the difference in luminance and color values between different pixels within a given image. Fig.~\ref{fig:contrast_example} shows an image taken from the ImageNet dataset at three different contrast settings. The original image is in the center, the image at half the original contrast value is on the left and the image at twice the original contrast value is on the right. Although there are many definitions of image contrast, such definitions generally follow the form:
$$\frac{\Delta \text{luminance}}{\text{avg. luminance}}$$
High contrast images have their pixel values spread out over a greater range of values, while low contrast images have a narrow range of pixel values. As a result, details in high contrast images are more visible than those in low contrast images. Intuitively, this ``sharpening'' effect on details within an image should lead to higher classification accuracy — as image details are more visible, a classifier can more easily identify and use such features for its prediction. However, Dodge et al. \cite{dodge2016understanding} discovered that image contrast appears to have little effect on classifier performance. When they reduced the image contrast of input images by blending the original images with gray images, performance degrades only if contrast is significantly reduced. As previous work has not found that image contrast affects network classification substantially, image contrast is not widely considered as an important data augmentation step. However, as we later demonstrate in our experiments, we discover that image contrast indeed plays an important role in adversarial attack and compression defense performance. 

\section{Experimental Results}
\label{sec:experiments}

In this section, we study the effects of input size and image contrast on adversarial machine learning using five popular image classification datasets: MNIST \cite{mnist}, Fashion-MNIST \cite{fmnist}, CIFAR10 \cite{cifar}, CIFAR100 \cite{cifar}, and ImageNet \cite{imagenet}.  The MNIST \cite{mnist} dataset consists of 70,000 $32 \times 32$ grayscale images split into 10 classes for handwritten digit recognition. The Fashion-MNIST dataset consists of 70,000 $32 \times 32$ grayscale images split into 10 different classes of fashion items. As Fashion-MNIST images are of fashion items rather than handwritten digits, they appear more visually complex than MNIST images. The CIFAR10 and CIFAR100 \cite{cifar} consists of 60,000 $32 \times 32$ color images split into 10 and 100 classes respectively. Finally, the ImageNet image classification dataset \cite{imagenet} consists of 1.2 million images of drawn from 1000 classes.

On each benchmark we evaluate four different white box attacks: FGSM \cite{goodfellow2014explaining}, BIM \cite{kurakin2016adversarial}, PGD \cite{madry2017deep}, and DeepFool \cite{moosavidezfooli2015deepfool}. We use the $L_{\infty}$ norm and $\epsilon \in \{\frac{2}{255}, \frac{4}{255}, \frac{8}{255}, \frac{16}{255}\}$ for all attacks. On each benchmark, we also evaluate three defense settings: no defense, TVM compression defense \cite{guo2017countering}, and JPEG compression defense \cite{dziugaite2016study}. We use four NVIDIA Tesla V100 GPUs for all of our experiments.

\subsection{Input Size Experiments}

To study the effect of image input size on the performance of adversarial attack and defense methods, we use three different input image sizes for each of our datasets and rescale the original images to the given input size using a bilinear interpolation. For MNIST, Fashion-MNIST, CIFAR10, and CIFAR100, we evaluate on three separate image sizes: half of the original input size, the original input size, and double the original input size. As ImageNet images are significantly larger than images from the other datasets, we evaluate on ImageNet images scaled to a fourth of the original size, half the original size, and the original images.

\begin{figure*}[h]
\centering
\includegraphics[scale=0.25]{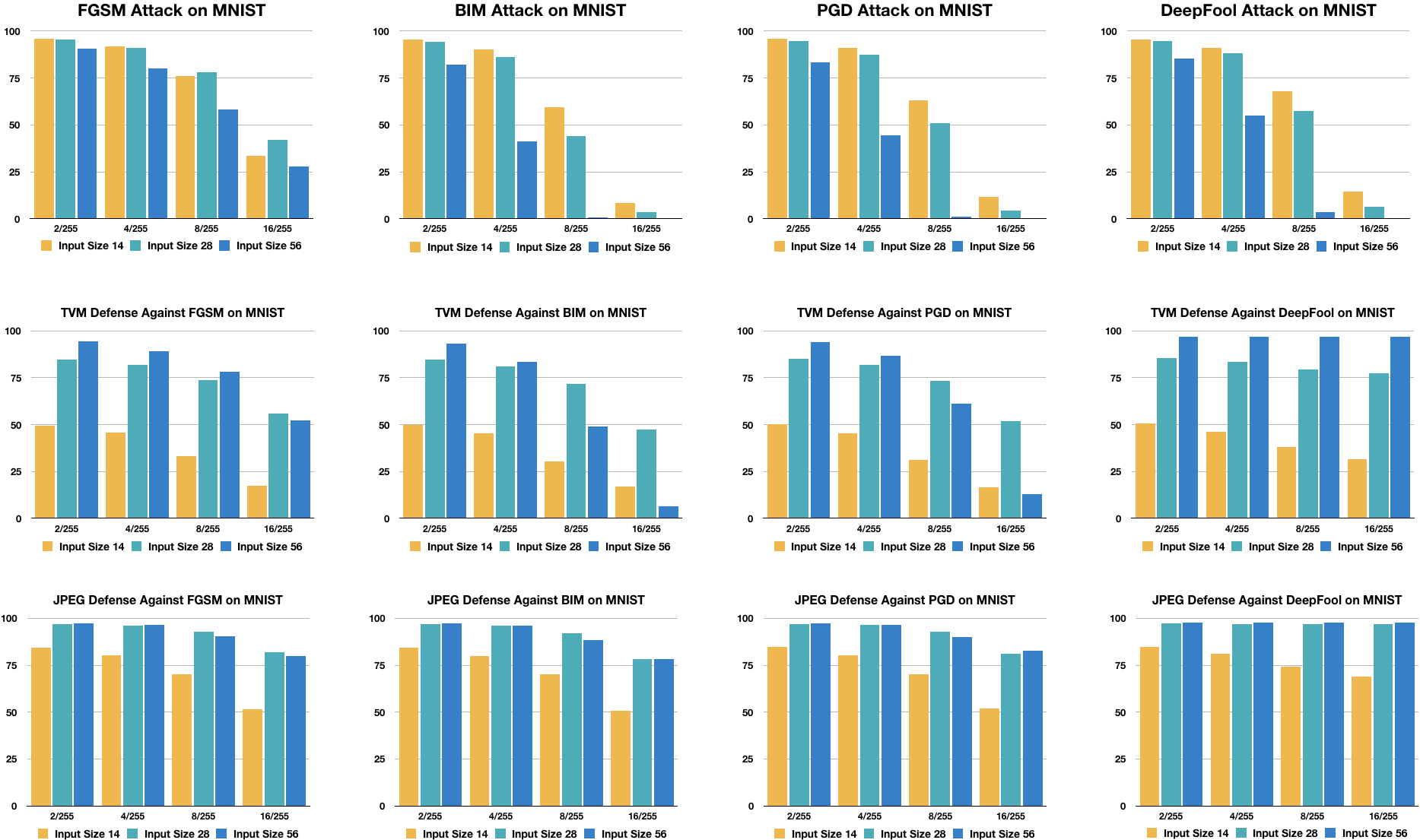} \\
\caption{
Input size results on MNIST. Each row represents a given defense (no defense, TVM, and JPEG) while each column represents a given attack (FGSM, BIM, PGD, DeepFool). Four different $\epsilon$ values are used in every attack-defense combination: 2/255, 4/255, 8/255, and 16/255.}
\label{fig:mnist_input_size} 
\end{figure*}

 On all datasets, we apply random resized crops and random horizontal flipping as data augmentation steps to boost classifier accuracy during training. Additionally, we also apply random rotations of up to 15 degrees to CIFAR100 images as an additional augmentation step to boost the network accuracy to state of the art.

\subsubsection{Experiments on MNIST/Fashion-MNIST}\hfill\par
\textbf{Settings:}
As we use images scaled to both half and double the original input size for our experiments, we additionally evaluate on $14 \times 14$ and $56 \times 56$ images. For both datasets, we adapt the convolutional neural network described by Madry et. al \cite{madry2017deep}, consisting of two convolutional layers with 32 and 64 filters respectively, each followed by a pooling layer, and a fully connected layer of size 1024. Instead of using $2 \times 2$ max-pooling after the second convolutional layer, we instead use an adaptive max-pooling layer to support dynamic input shapes for our experiments. For training, we use the Adam optimizer with a learning rate of $0.01$ and betas of $0.9$ and $0.999$. We run our model for 100 epochs using a batch size of 64, using PyTorch's learning rate scheduler to reduce the learning rate if no improvement is seen in the loss after five consecutive epochs.

\begin{figure*}[h]
\centering
\includegraphics[scale=0.25]{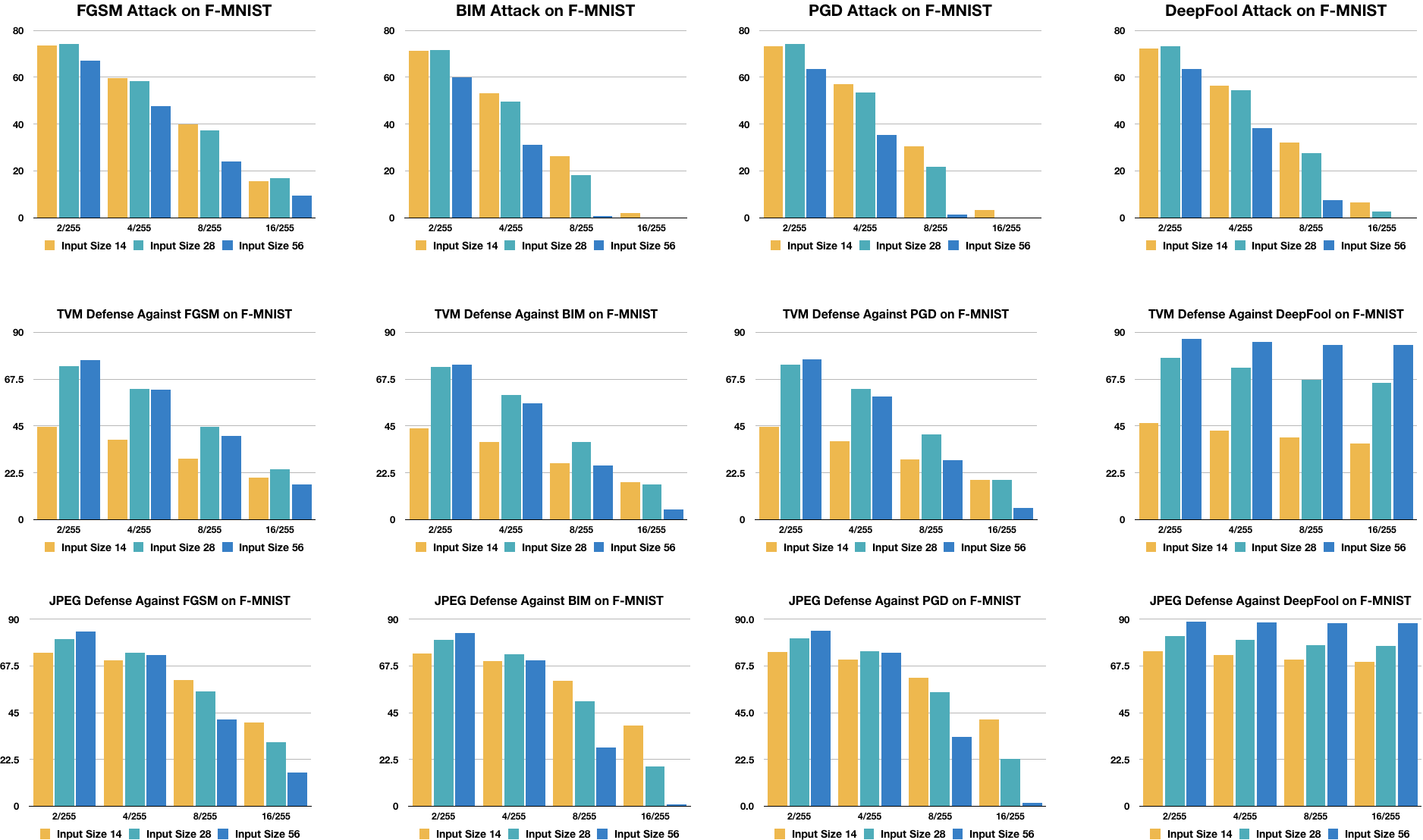} \\
\caption{
Input size results on Fashion-MNIST. Each row represents a given defense (no defense, TVM, and JPEG) while each column represents a given attack (FGSM, BIM, PGD, DeepFool). Four different $\epsilon$ values are used in every attack-defense combination: 2/255, 4/255, 8/255, and 16/255.}
\label{fig:fmnist_input_size} 
\end{figure*}

\textbf{Results:}
The results of the adversarial attacks and defense methods are shown in Fig.~\ref{fig:mnist_input_size} for MNIST and in Fig.~\ref{fig:fmnist_input_size} for Fashion-MNIST. Our original MNIST model achieves a top-1 accuracy of 98.39\%, while our Fashion-MNIST model achieves a top-1 accuracy of 91.20\%. We discover that given attacks and defenses perform differently as a result of the input image size. On both datasets and across all attacks, input size correlates negatively with adversarial robustness when no defense is applied — $14 \times 14$ images consistently perform better against attacks with only one outlier. We believe this may result from overfitting as both MNIST and F-MNIST are fairly simple datasets, potentially resulting in the CNN overfitting and becoming vulnerable to adversarial exploitation. By training on smaller input images with less features, we force the CNN to learn a better underlying representation of the data, making it more robust to adversarial examples. Another explanation may be that smaller input images limit an attacker's choice of possible perturbations simply because smaller images contain fewer pixels to modify. 

However, when TVM and JPEG defenses are applied to these attacks, larger images tend to lead to better adversarial robustness. In fact, for $\epsilon \in \{\frac{2}{255}, \frac{4}{255}, \frac{8}{255}\}$ across all attacks, the TVM defense does more damage than the original attack for $14 \times 14$ images. Compression defenses like TVM and JPEG may disproportionately impact smaller input images by destroying useful features vital for classification, leading to worse defense performance compared to larger images, where compression will leave behind more features while still smoothing adversarial perturbations. For small $\epsilon$ values, the boost in defense performance from using larger input images outweighs the boost in attack performance from larger images, as the attack method is limited by the small value of $\epsilon$.

For large values of $\epsilon$, however, the increased ease of attack may outweigh the benefits that a large input size offers to compression defense approaches: the massive drop in the accuracy of $56 \times 56$ images in Fig.~\ref{fig:mnist_input_size} (TVM-BIM and TVM-PGD combinations) and Fig.~\ref{fig:fmnist_input_size} (all TVM and JPEG defenses except on DeepFool) shows that the larger input size gives the attack more possible perturbations, allowing it to find a better adversarial example. Intuitively, adversarial attack is easier than defense as an attack only needs to perturb clean images, while a defense must successfully defend against adversarial examples while still allowing the model to classify clean examples correctly. Thus in these cases, $14 \times 14$ images perform much better than $56 \times 56$ images for $\epsilon=\frac{16}{255}$. This is very interesting, as smaller images can be more robust under adversarial attacks in both situations, without any defense and with compression defense. 

 



We note that this trend does not hold true for DeepFool. As shown in the last column of Fig.~\ref{fig:mnist_input_size} and Fig.~\ref{fig:fmnist_input_size}, the defense performance across input sizes is roughly the same regardless of $\epsilon$ value. Even for the largest $\epsilon$ value of $\frac{16}{255}$, both the TVM and JPEG defenses return the classifier accuracy to more than 97\% for $56 \times 56$, close to the original classification accuracy. This indicates that both TVM and JPEG are performing significantly better against the DeepFool attack than against other attacks. Additionally, given that both defenses appear to reach virtually the same defended accuracy regardless of $\epsilon$ value, we believe that both defenses are performing very well for the DeepFool attack, suggesting that compression-based defenses successfully remove adversarial perturbations generated by DeepFool. We hypothesize the higher accuracy of defense methods on DeepFool compared to our other attack methods lies in the difference between DeepFool and our other attack methods. Instead of perturbing in the direction that maximizes the cost function, DeepFool perturbs in the direction orthogonal to the linearized decision boundary of the attacked model.


\subsubsection{Experiments on CIFAR10/CIFAR100}\hfill\par
\textbf{Settings:}
For our experiment, we use input images sizes $\in \{16 \times 16, \:32 \times 32, \:64 \times 64\}$. For both datasets, we use the ResNet-18 architecture layout specified for CIFAR by He et al. \cite{resnet} with one modification, changing the average pooling layer before the fully connected layer to an adaptive average pooling layer to support dynamic input sizes.

For training CIFAR10, we use SGD with momentum, setting an initial learning rate of $0.1$, weight decay of $5e^{-4}$, and momentum of $0.9$. We run our model for 350 epochs using a batch size of 128, scaling the initial learning rate by $0.1$ at epochs 150 and 250. On CIFAR100, we use SGD with momentum, setting an initial learning rate of $0.05$, weight decay of $0.001$, and momentum of $0.9$. We run our model for 200 epochs using a batch size of 128, scaling the initial learning rate by $0.2$ at epochs 60, 120, and 160.

\begin{figure*}[ht!]
\centering
\includegraphics[scale=0.25]{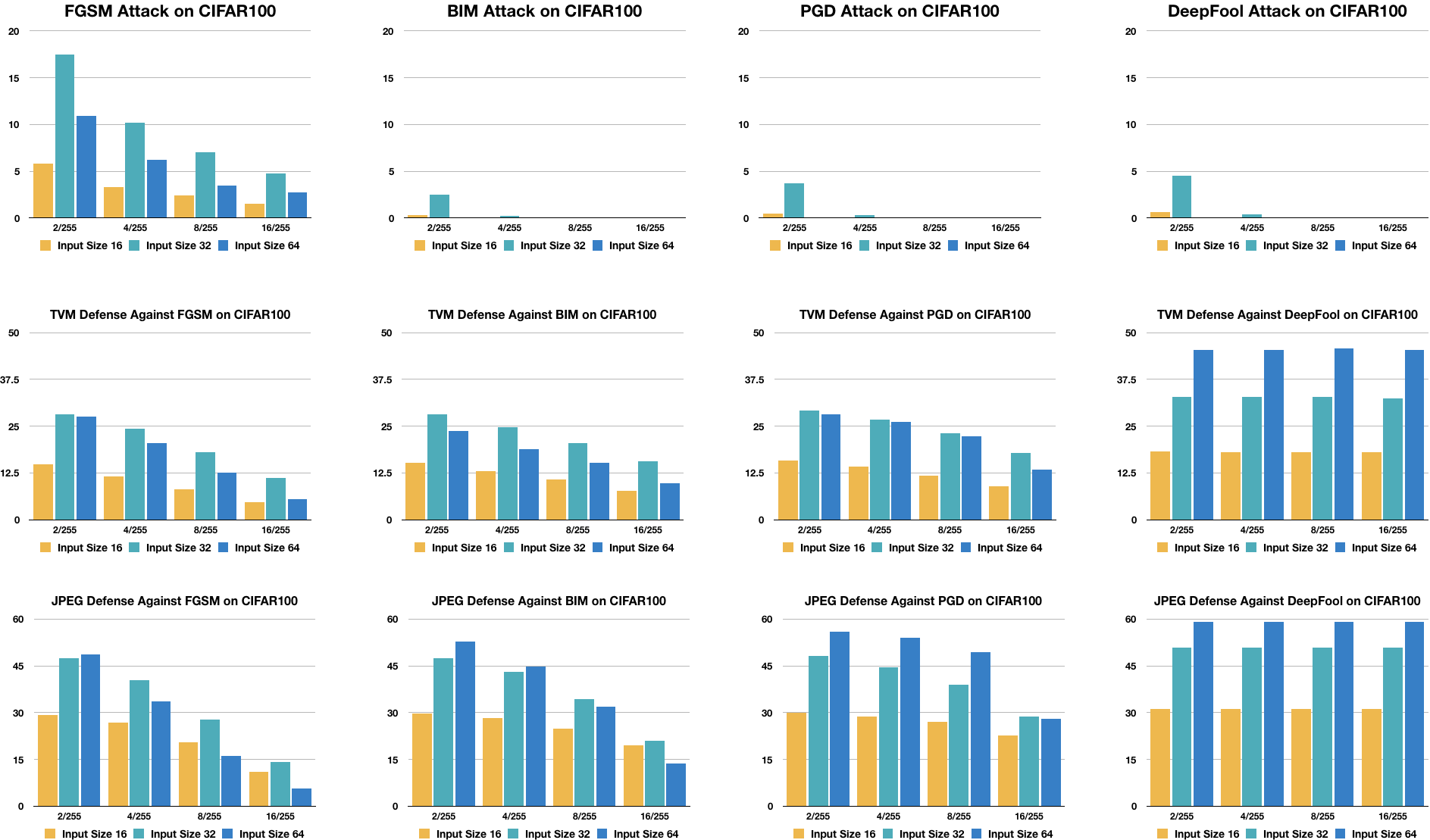} \\
\caption{
Input size results on CIFAR100. Each row represents a given defense (no defense, TVM, and JPEG) while each column represents a given attack (FGSM, BIM, PGD, DeepFool). Four different $\epsilon$ values are used in every attack-defense combination: 2/255, 4/255, 8/255, and 16/255.}
\label{fig:cifar100_input_size} 
\end{figure*}

\textbf{CIFAR10 Results:}
Our original CIFAR10 model achieves a top-1 accuracy of 94.77\%. However, we ran into difficulty training models from scratch on our resized images — despite trying multiple hyperparameter settings\footnote{We tried training with both our original CIFAR10 and CIFAR100 hyperparameters, training with initial learning rates $\in \{0.1, 0.05, 0.01, 0.001\}$ with scaling factors $\in \{0.1, 0.2\}$, and also a learning rate scheduler to reduce the learning rate every five consecutive epochs with no improvement. However, none of these approaches were able to successfully converge.} with both SGD and Adam optimizers, we were unable to get a trained model above 65\% accuracy without finetuning from the weights trained on the $32 \times 32$ images. We suspect by training using the $32 \times 32$ image weights, our classifiers for the reshaped images may have underfit the new reshaped images, with most weights remaining similar to the original weights trained on $32 \times 32$ images. As Goodfellow et al. \cite{goodfellow2014explaining} notes, such underfitting generally results in a model more vulnerable to adversarial attack. However, despite the increased adversarial vulnerability of our models trained on reshaped inputs, we still see similar trends to CIFAR100. Thus, due to space contraints, we only show our results on CIFAR100 and provide all of our graphs, including our CIFAR10 results, on our GitHub \cite{jzpan_adversarial-playground}.

\textbf{CIFAR100 Results:}
The results of the adversarial attacks and defense methods are shown in Fig.~\ref{fig:cifar100_input_size}. Our original CIFAR100 model achieves a top-1 accuracy of 74.57\%. In contrast to MNIST and Fashion-MNIST, we discover that when no defense is applied, the original input size of $32 \times 32$ performs best across all attacks and $\epsilon$ values. We conjecture that the low classification accuracy on CIFAR100 for $16 \times 16$ images stems from the nature of the classification task. CIFAR100 is a significantly harder classification task than CIFAR10 as the images are split into more classes (100 vs 10 classes). Additionally, CIFAR100 classes are fairly similar to each other — the dataset defines groups of five classes into ``superclasses" (e.g. the beaver, dolphin, otter, seal, and whale classes belong to the ``aquatic mammals" superclass). As CIFAR100 only has 500 small ($32 \times 32$) images per class, a well-trained classifier on CIFAR100 must classify a low-quality image into one of 100 fairly similar classes with very little training data. By downsampling CIFAR100 images to $16 \times 16$ and squeezing potentially useful features, we only make it more difficult for our network to learn a good underlying representation of the data and thus increase the model's susceptibility to adversarial examples. 

Additionally, the upsampled CIFAR100 images become more vulnerable to adversarial attacks as the increased input size allows the attack method to search a larger space of possible perturbations. $32\times 32$ images thus lie in a ``sweet spot" where the images are large enough for the classifier to learn a good underlying representation of the data, but not large enough to boost the performance of adversarial attack methods. Thus, in scenarios where no defense is applied, using the smallest imput image size that still allows the model to learn well may increase a model's robustness to adversarial attack. This is consistent with our observation in MNIST/Fashion-MNIST.

When the TVM defense is applied, we also see that $32 \times 32$ images are more robust in almost every case.  $32 \times 32$ images perform 14\% better than $16 \times 16$ images and 2-5\% better than $64 \times 64$ images for small $\epsilon$. For large $\epsilon$, $32 \times 32$ images perform 5\% better than their resized counterparts. Again, this suggests that $32\times 32$ images thus lie in a ``sweet spot" that allows for accurate classification without compromising adversarial robustness.

When the JPEG defense is applied, we see that CIFAR100 is consistent with our previous experiments. As with MNIST/Fashion-MNIST, the input image size on CIFAR100 correlates positively with adversarial robustness for small $\epsilon$ but defense performance drops faster for larger input sizes as $\epsilon$ increases. In the JPEG-defense-against-BIM case, we again see that the smallest image size is the most robust for $\epsilon=16/255$.  
We also notice that the smallest input size has the slowest drop in accuracy as $\epsilon$ increases. This suggests that the limitation on the classification accuracy of $16 \times 16$ images may be due to model learning a poor representation rather than larger values of $\epsilon$ increasing attack strength.

Across both TVM and JPEG, we notice the same exception with our results against the DeepFool attack, i.e., the defense performance against the attack remains virtually the same regardless of $\epsilon$, as shown in the last column of Fig.~\ref{fig:cifar100_input_size}.

\begin{figure*}[ht!]
\centering
\includegraphics[scale=0.25]{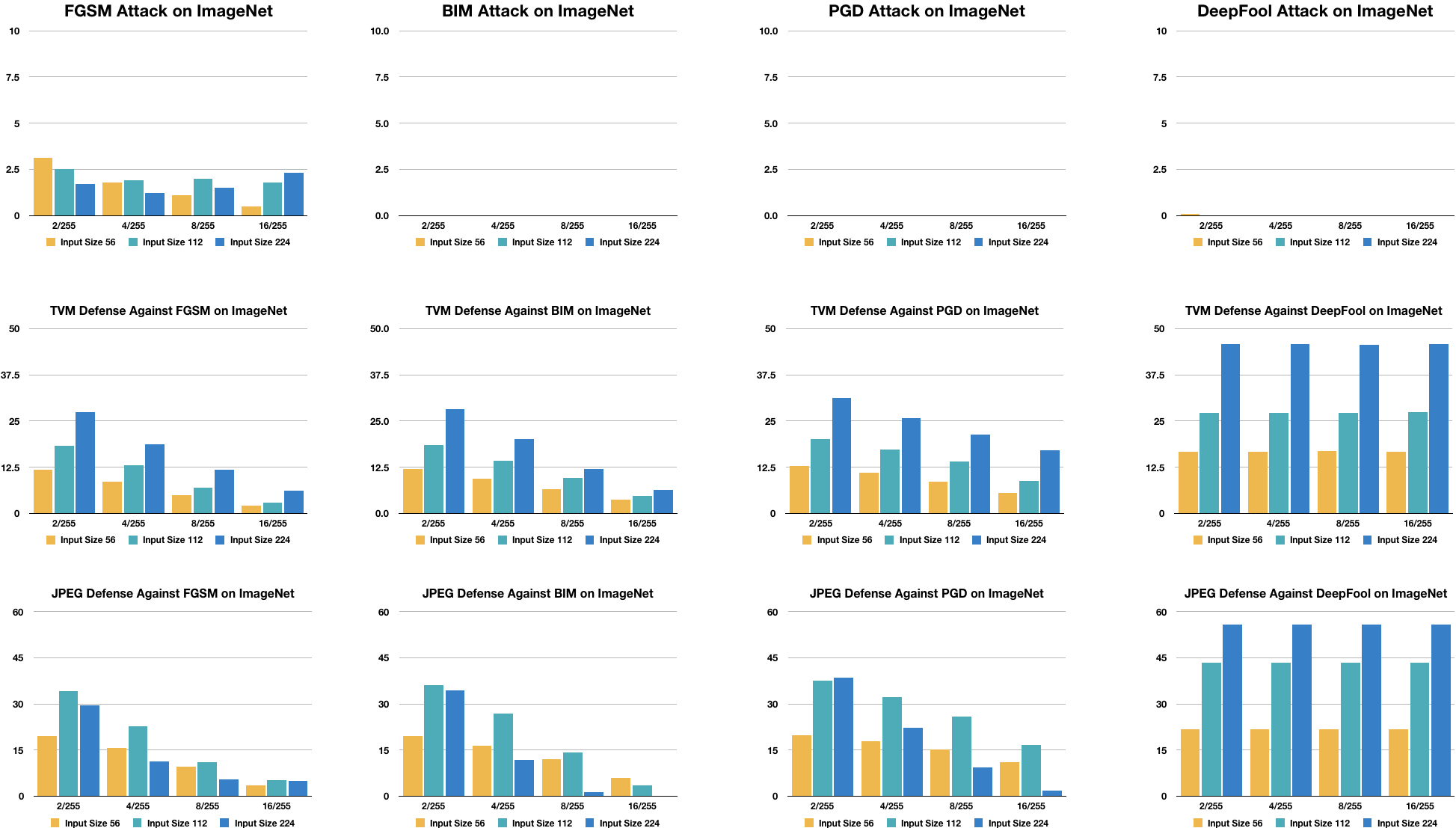} \\
\caption{
Input size results on ImageNet. Each row represents a given defense (no defense, TVM, and JPEG) while each column represents a given attack (FGSM, BIM, PGD, DeepFool). Four different $\epsilon$ values are used in every attack-defense combination: 2/255, 4/255, 8/255, and 16/255.}
\label{fig:imagenet_input_size} 
\end{figure*}

\subsubsection{Experiments on ImageNet}\hfill\par
\textbf{Settings:}
We resize ImageNet input images to $\{56 \times 56, \:112 \times 112, \:224 \times 224\}$. We use a standard ResNet-18 architecture with an adaptive average pooling layer before the fully connected layer, trained with the SGD optimizer with an initial learning rate of 0.1, weight decay of $1e^{-4}$, and momentum of $0.9$. We run our model for 90 epochs using a batch size of 256, scaling the initial learning rate by $0.1$ at epochs 30 and 60.

\textbf{Results:}
The results of the adversarial attacks and defense methods are shown in Fig.~\ref{fig:imagenet_input_size}. Our original ImageNet model achieves a top-1 accuracy of 70.12\%. As all adversarial attacks perform very well on ImageNet when no defense is applied, with accuracy ranging from 0 to 3.1\%, we do not try to extrapolate any trends from such results. As our attacks are based off of first-order optimization, any accuracy differences across different input sizes could result from the stochastic nature of each attack method. 

However, we notice consistent trends across input sizes once we apply the TVM and JPEG defenses. Again, as with the previously tested datasets, input size correlates positively with adversarial robustness for small $\epsilon$ but defense performance drops steeply for larger input sizes at larger $\epsilon$. For the TVM defense, we see the defense performance on $224 \times 224$ drop steeper than for other input sizes, though TVM for $224 \times 224$ images still claims the highest defended accuracy across all $\epsilon$ values. For the JPEG defense, we see the tradeoff between increased ease of attack and increase ease of defense fully swing the opposite way as $224 \times 224$ images perform very poorly for values of $\epsilon$ $\in \{\frac{8}{255}, \frac{16}{255}\}$. The difference in trends between the TVM and JPEG defenses suggests that the TVM defense benefits much more from a larger input size than the JPEG defense.

Across both TVM and JPEG, we again notice the same exception with our results against the DeepFool attack, i.e., the defense performance against the attack remains virtually the same regardless of $\epsilon$, as shown in the last column of Fig.~\ref{fig:imagenet_input_size}. Both defenses are very effective, in particular on larger image sizes.

\subsection{Image Contrast Experiments}

We next study the effect of image contrast on the performance of adversarial attack and defense methods. For each dataset we pick three different contrast levels: half the original contrast value, the original contrast value, and double the original contrast value. We retrain a new network for each dataset and contrast setting using the settings specified in our input size experiments.

\subsubsection{Experiments on MNIST/Fashion-MNIST}

\begin{figure*}[ht!]
\centering
\includegraphics[scale=0.25]{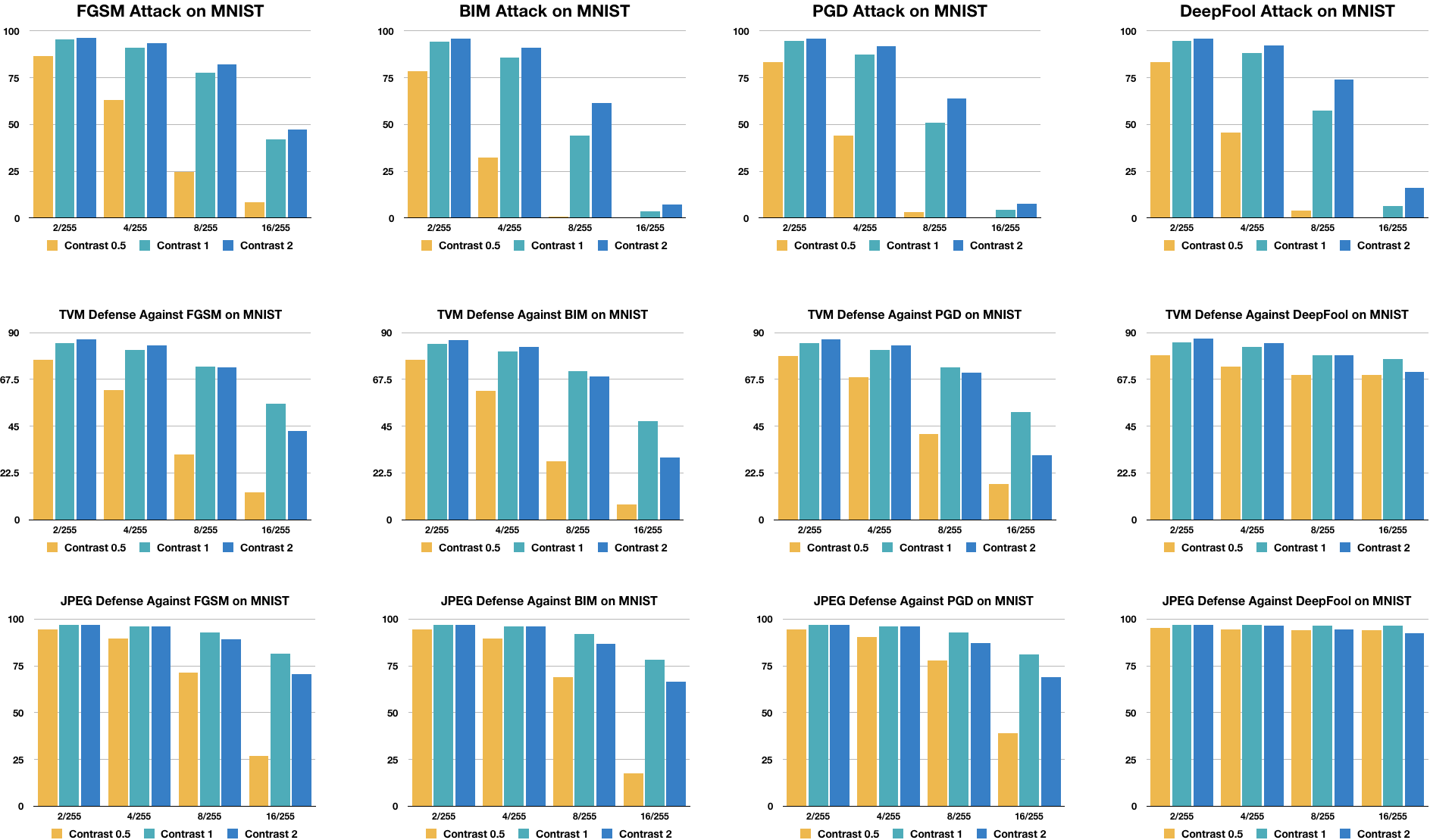} \\
\caption{
Contrast results on MNIST. Each row represents a given defense (no defense, TVM, and JPEG) while each column represents a given attack (FGSM, BIM, PGD, DeepFool). Four different $\epsilon$ values are used in every attack-defense combination: 2/255, 4/255, 8/255, and 16/255.}
\label{fig:mnist_contrast} 
\end{figure*}

The results of the adversarial attacks and defense methods are shown in Fig.~\ref{fig:mnist_contrast}. We only display MNIST results to save space, since F-MNIST and MNIST results are very similar, and the complete graphs are posted on our GitHub \cite{jzpan_adversarial-playground}. Our three trained models, each trained on a different contrast setting, reach roughly the same original classification accuracy. Our $0.5 \times$ contrast model reaches 97.97\% accuracy, our original contrast model reaches 98.34\% accuracy, and our $2 \times$ contrast model reaches 98.44\% accuracy. As Fig.~\ref{fig:mnist_contrast} shows, higher contrast images appear to be more adversarially robust in all three cases (no defense, TVM, and JPEG) and across attack methods. The accuracy on the reduced contrast setting consistently trails the accuracies on the other contrast settings by around 7-8\% for $\epsilon = \frac{2}{255}$ when no defense is applied. The accuracy gap between the normal contrast setting and the increased contrast setting is smaller but still noticeable: around 2-4\%.

By increasing the image contrast, we make recognizing defining features easier for the network: edges and other features important for correct image classification become more visible. We believe this increased ease of classification may make it more difficult for adversarial attacks to successfully cause the attacked network to misclassify. However, we notice that for larger $\epsilon$ settings, the accuracy of the classifier on modified contrast images degrades much more steeply than for the original images. The gap between the lower contrast setting and the other settings for the FGSM attack with $\epsilon = \frac{16}{255}$ increases to 33\%, a massive performance gap. In several cases when $\epsilon = \frac{2}{255}$ (such as when defenses are applied against FGSM), the accuracy for the $2 \times$ contrast images falls beneath the accuracy of original contrast images. As with input image size, we believe that finding an appropriate tradeoff is important — although images with very little contrast are difficult to classify, increasing image contrast too much can also destroy features that a DNN relies upon for classification. Images with too much contrast may skew the training data, making it more difficult for a DNN to learn a distribution that fits the underlying data. This only increases the adversarial vulnerability of the DNN. Although a model's weak representation may result in higher clean image classification and remains resilient to adversarial examples when $\epsilon$ is small, the model's weaknesses become apparent once the value of $\epsilon$ increases — for larger $\epsilon$ values across attack methods, performance degrades significantly faster for the modified contrast images. Our experiment demonstrates the importance of finding an appropriate contrast setting for input images as a data preprocessing step — a good contrast setting can boost the performance of defense methods significantly in some cases, boosting the performance of the JPEG defense against BIM by more than 50\% in the best case scenario.

\begin{figure*}[ht!]
\centering
\includegraphics[scale=0.25]{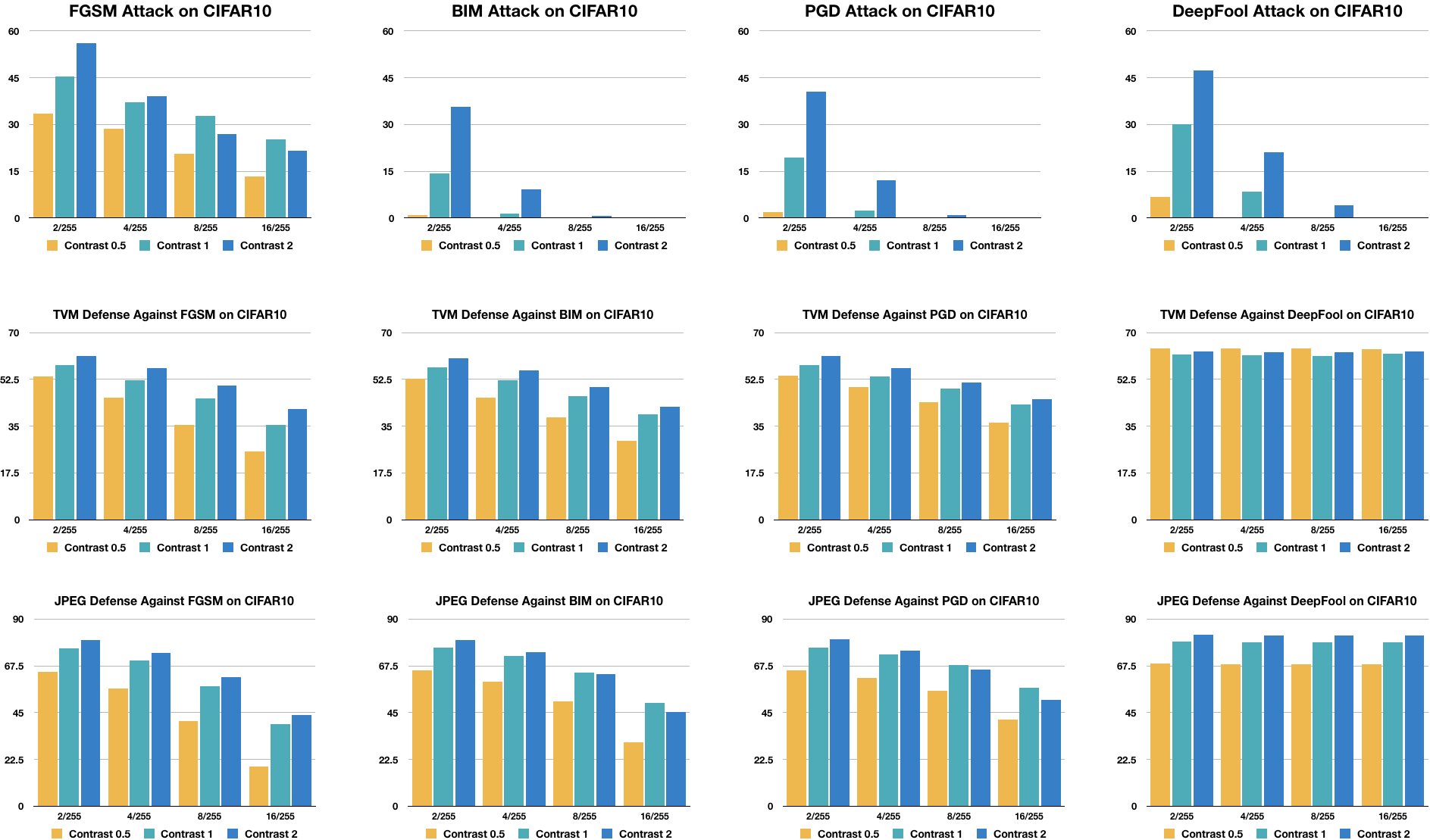} \\
\caption{
Contrast results on CIFAR10. Each row represents a given defense (no defense, TVM, and JPEG) while each column represents a given attack (FGSM, BIM, PGD, DeepFool). Four different $\epsilon$ values are used in every attack-defense combination: 2/255, 4/255, 8/255, and 16/255.}
\label{fig:cifar10_contrast} 
\end{figure*}

\subsubsection{Experiments on CIFAR10/CIFAR100}

The results of the adversarial attacks and defenses are shown in Fig.~\ref{fig:cifar10_contrast}. We only display CIFAR10 results to save space, since CIFAR10 and CIFAR100 results are very similar, and the complete graphs are posted on our GitHub \cite{jzpan_adversarial-playground}. Again our three models reach roughly the same original classification accuracy. On CIFAR10, our $0.5 \times$ contrast model reaches 94.18\% accuracy, our original contrast model reaches 94.77\% accuracy, and our $2 \times$ contrast model reaches 93.10\% accuracy. On CIFAR100, our $0.5 \times$ contrast model reaches 75.13\% accuracy, our original contrast model reaches 74.57\% accuracy, and our $2 \times$ contrast model reaches 72.15\% accuracy. As with MNIST/Fashion-MNIST, CIFAR10/CIFAR100 also demonstrates that adversarial robustness mostly increases with higher contrast images. However, we again note that in several cases when $\epsilon = \frac{16}{255}$, the $2 \times$ contrast images perform worse than the original images. This again suggests that although increasing image contrast boosts adversarial robustness, too high of an image contrast setting may harm the classifier's ability to generalize, making it susceptible to adversarial examples. Finally, defended accuracy on DeepFool again appears consistent across $\epsilon$ values, suggesting that the tested compression defenses perform very well.

\subsubsection{Experiments on ImageNet}

\begin{figure*}[ht!]
\centering
\includegraphics[scale=0.25]{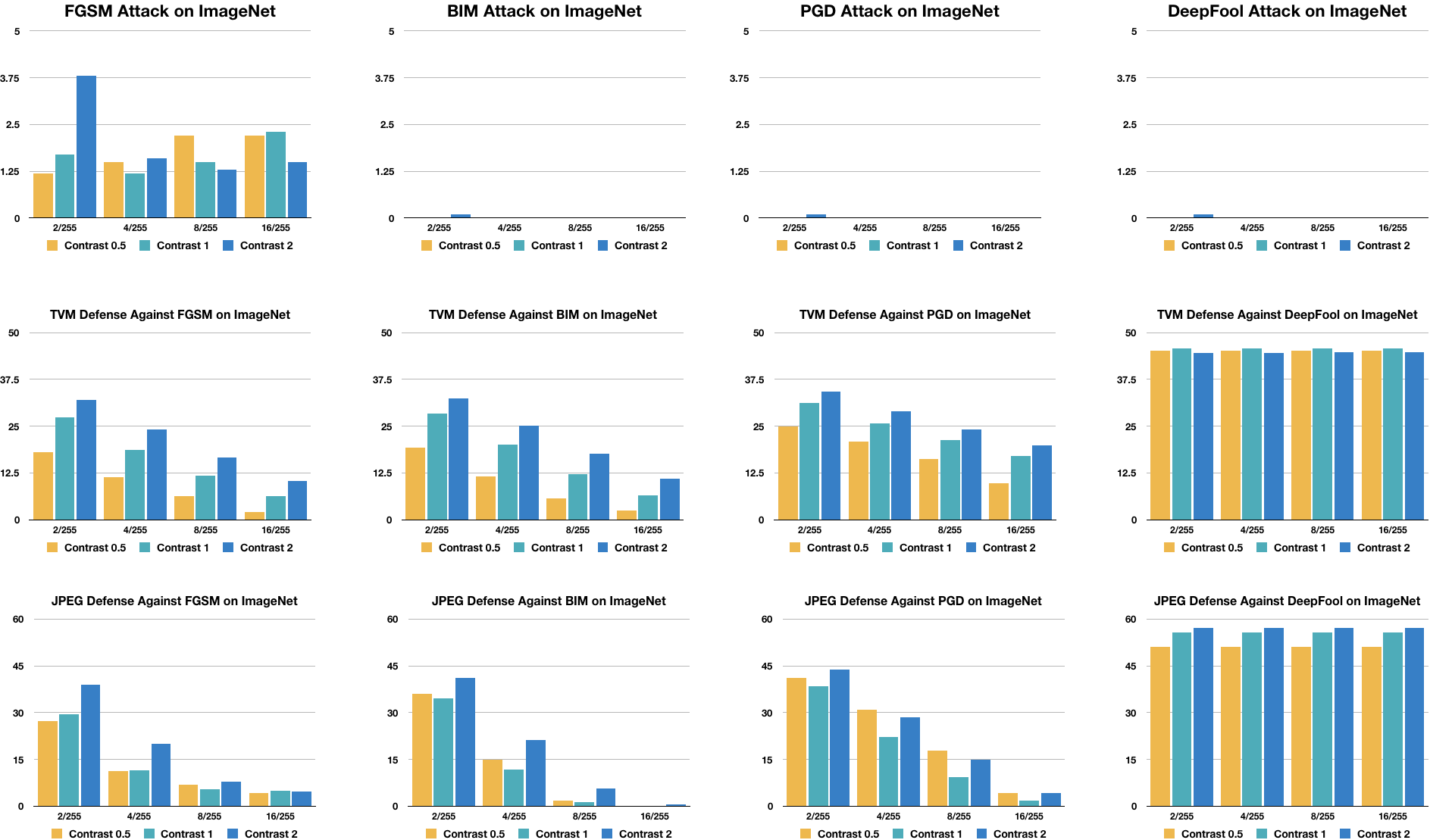} \\
\caption{
Contrast results on ImageNet. Each row represents a given defense (no defense, TVM, and JPEG) while each column represents a given attack (FGSM, BIM, PGD, DeepFool). Four different $\epsilon$ values are used in every attack-defense combination: 2/255, 4/255, 8/255, and 16/255.}
\label{fig:imagenet_contrast} 
\end{figure*}

The results on ImageNet are shown in Fig.~\ref{fig:imagenet_contrast}. Again our three models reach roughly the same original classification accuracy. Our $0.5 \times$ contrast model reaches 69.54\% accuracy, our original contrast model reaches 70.12\% accuracy, and our $2 \times$ contrast model reaches 67.98\% accuracy. Given the effective performance of our attacks on ImageNet when no defense is applied, reducing the accuracy to below 5\% in all cases, we choose not to draw any conclusions, as the stochastic nature of training and adversarial example generation could account for any trends we may try to draw. Looking at the performance of the TVM defense, we see that ImageNet very clearly follows the trend of larger contrast settings achieving higher accuracy on adversarial examples across all attacks and $\epsilon$ settings. Intriguingly, we notice that on BIM and PGD, the more advanced cost-function-based attacks, the JPEG defense actually performs the worst for the original contrast setting. The defended accuracy against DeepFool again is consistent across $\epsilon$ values, suggesting that the tested compression defenses perform very well.

\section{Conclusion and Future Direction}
\label{sec:conclusion}

In this paper, we investigate intrinsic dataset properties on adversarial machine learning. In particular, we focus on two intrinsic image dataset properties, namely image size and image contrast, and test our hypothesis using four different attacks and two different defense techniques on five popular image classification datasets with different maximum allowable perturbations. We have found a number of interesting observations. 

\begin{itemize}
\item Image size and image contrast play important roles in the effectiveness of adversarial attacks and defenses.

\item In general, smaller image sizes are more robust to adversarial attacks, but when the compression defense techniques TVM and JPEG are applied, larger image sizes can counter such attacks more effectively. However, for some attack-defense-dataset combinations (e.g., BIM-JPEG for Fashion-MNIST and ImageNet with large $\epsilon$ settings), the smallest image sizes are also more robust with defense. This could have very interesting applications to future edge applications of AI where small image sizes are strongly preferred due to computing and communication constraints. Our study shows that smaller images can actually improve the adversarial robustness of DNNs.

\item In terms of image contrast, higher image contrast usually leads to better adversarial robustness, with or without defenses. There are also some exceptions where too large of a contrast setting may backfire (e.g., PGD-JPEG for CIFAR10 with $\epsilon = 16/255$). More detailed studies are needed to pinpoint the reasons behind this decrease in robustness and find the best image contrast for adversarial robustness when new datasets are designed.

\item DeepFool, although an effective attack technique, can be easily defended, across different image sizes and contrasts, for all five datasets and four $\epsilon$ settings.

\end{itemize}

Since this work, to our best knowledge, is the first systematic effort toward understanding the impacts of intrinsic dataset properties on adversarial machine learning, there are many open research problems and applications to explore. More dataset properties such as the presence of outliers and class imbalance can be studied. Additionally, jointly exploring dataset properties and ML task complexity may better explain why intrinsic dataset properties sometimes display different trends on different datasets. Even for the input image size and contrast problems that we study in this work, there are still unanswered questions, e.g., given an original dataset, how to augment or modify it such that the dataset will achieve the best adversarial robustness, with or without defenses. Such dataset preprocessing requires little computational overhead. Dataset preprocessing is inherently model agnostic, making it a cheap yet effective addition to any machine learning pipeline. However, as we show preprocessing techniques are sometimes attack/defense dependent, the optimal data preprocessing may depend on the deployment scenario. Well-designed datasets and preprocessing techniques in conjunction with existing defense methods can create more robust machine learning pipelines.

\section*{Acknowledgement}
\label{sec:acknowledgement}

The authors are grateful for the generous support of Dr. John Cohn of the MIT-IBM Watson AI Lab and Dr. Chris Hill of the Massachusetts Green High Performance Computing Center (MGHPCC) for providing access to the Satori computing cluster used to carry out this research. 

{
\bibliographystyle{IEEEtran}
\bibliography{ref/attacks,ref/defenses,ref/Top_sim,ref/ML,ref/background, ref/experiments}
}

\end{document}